\documentclass[runningheads]{llncs}
\usepackage{graphicx}
\usepackage[square,sort,comma,numbers]{natbib}
\usepackage{multirow}
\usepackage{booktabs}
\usepackage{xcolor}
\usepackage{hyperref}
\usepackage{amsmath}
\usepackage{amsfonts}
\usepackage[normalem]{ulem}
\usepackage{hyphenat}
\usepackage{subfig}
\usepackage{subcaption}
\usepackage{algorithm}
\usepackage{algpseudocode}
\usepackage{nicematrix}

\newcommand{\repo}{\url{https://github.com/j2launay/ebbwbb}}

\NiceMatrixOptions{code-for-first-row = \color{red},                   code-for-first-col = \color{magenta},                   code-for-last-row = \color{green},                   code-for-last-col = \color{blue}} 

\begin{document}
\title{Does It Make Sense to Explain a Black Box With Another Black Box?}
%
%
\author{Julien Delaunay\inst{1} 
\and Luis Galárraga\inst{1} \and Christine Largouët\inst{2}}
\authorrunning{J. Delaunay et al.}
%
\institute{INRIA/IRISA, Rennes, France \email{\{julien.delaunay, luis.galarraga\}@inria.fr} \and L’Institut Agro/IRISA, Rennes, France \email{christine.largouet@irisa.fr}} 
\maketitle              
\begin{abstract}
    Although counterfactual explanations are a popular approach to explain ML black-box classifiers, they are less widespread in NLP. 
    Most methods find those explanations by iteratively perturbing the target document until it is classified differently by the black box. We identify two main families of counterfactual explanation methods in the literature, namely, (a) \emph{transparent} methods that perturb the target by adding, removing, or replacing words, and (b) \emph{opaque} approaches that project the target document into a latent, non-interpretable space where the perturbation is carried out subsequently. This article offers a comparative study of the performance of these two families of methods on three classical NLP tasks. Our empirical evidence shows that opaque approaches can be an overkill for downstream applications such as fake news detection or sentiment analysis since they add an additional level of complexity with no significant performance gain. These observations motivate our discussion, which raises the question of whether it makes sense to explain a black box using another black box.\footnote{This article was originally published in French at the Journal TAL. VOL 64 n°3/2023}

\end{abstract}

\section{Introduction}
\label{sec:introduction}
The latest advances in machine learning (ML) have led to significant advances in various natural language processing (NLP) tasks~\cite{bert,roberta,DistilBERT}, such as text generation, fake news detection, sentiment analysis, and spam detection. These notable improvements can be partly attributed to the adoption of methods that encode and manipulate text data using latent representations. Those methods embed text into high-dimensional vector spaces that capture the underlying semantics and structure of language, and that are suitable for complex ML models. 

Despite the impressive gains in accuracy achieved by modern ML algorithms~\cite{bert,gpt3}, their utility can be diminished by their lack of interpretability~\cite{interpret_gan}. This has, in turn, raised an increasing interest in ML explainability, the task of providing appropriate explanations for the answers of black-box ML algorithms~\cite{trends_in_xai}. Indeed, a model could make correct predictions for the wrong reasons~\cite{bias_nlp,bias_nlp2}. Unless the ML model is a white box, explaining the results of such an agent requires an explanation layer that elucidates the internal workings of the black box in a post-hoc manner.

\begin{figure}[t]
    \includegraphics[width=\textwidth]{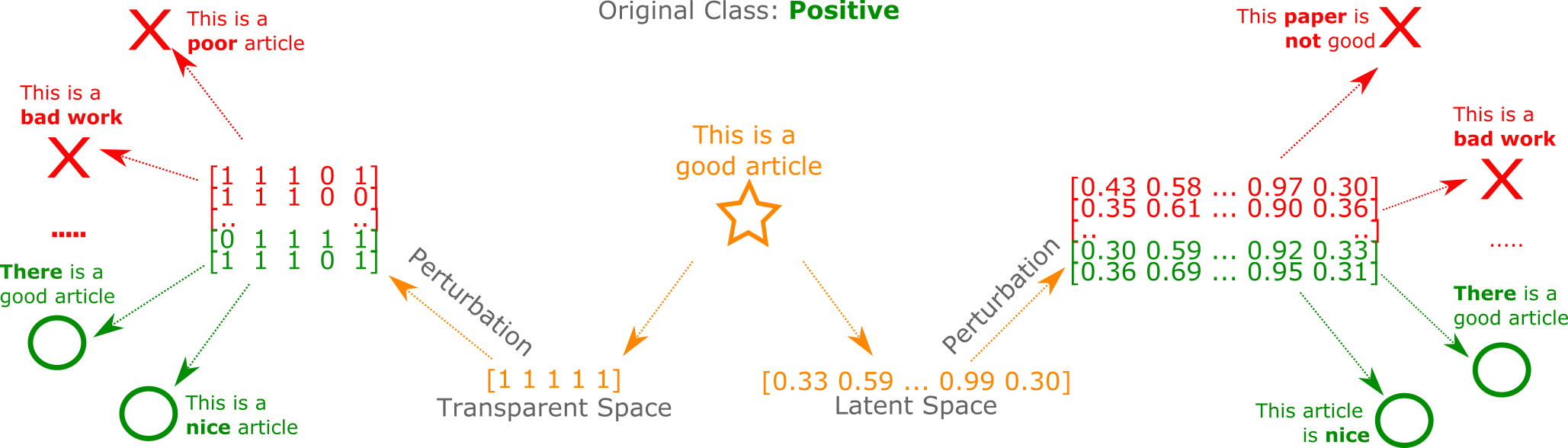}
    \caption{The mechanism employed to perturb the target documents by the transparent and opaque methods. Transparent techniques, on the left, convert the input text to a vector representation, where `1' indicates the presence of the input word and `0' denotes a replacement. Opaque methods, as on the right, embed words from the target text into a latent space and perturb the text in this high-dimensional space.}
    \label{fig:text_cf}
\end{figure}
  
While there are several ways to explain the outcomes of an ML model a posteriori, there has been a growing emphasis on counterfactual explanations, a domain that has experienced notable popularity over the last five years~\cite{survey_cf,Miller}. A counterfactual explanation is a counter-example that is similar to the original text, but that elicits a different outcome in the black box~\cite{wachter}. Consider the classifier depicted in Figure~\ref{fig:text_cf}, for sentiment analysis applied to the review ``This is a good article'' -- classified as positive. In this toy example, a counterfactual could be the phrase ``This is a \textbf{poor} article''. This explanation tells us that the adjective ``good'' was a possible reason for this sentence to be classified as positive, and changing the polarity of that adjective changes the classifier's response. 

In the literature, counterfactual explanation methods operate by increasingly perturbing the target text until the answer from the model -- often a classifier -- changes. Those perturbations can be conducted \emph{transparently} by adding, removing, or changing words and syntactic groups~\cite{martens,plausible_counterfactual,mice} in the original target text as depicted in Figure~\ref{fig:text_cf}. Since removing or adding words from a text can lead to unrealistic texts,
more recent methods~\cite{decision_boundary,cfgan,xspells} embed the target text in a latent space that captures the underlying distribution of the model's training corpus. Perturbations are then carried out in this space and then brought back to the space of words to guarantee realistic counterfactual explanations. These explanation methods rely on \emph{opaque} sophisticated techniques to compute those explanations~\cite{gan_interpret_latent}, which is tantamount to explaining a black box with another black box.

Based on this somehow paradoxical observation, we conduct a comparative study of various transparent and opaque post-hoc counterfactual explanation approaches. Rather than two distinct categories, the studied methods define a continuum, as some methods may combine transparent and non-interpretable techniques. Our empirical analyses have revealed that, for certain NLP tasks such as spam detection, fake news detection, or sentiment analysis, learning a compressed representation may be unnecessary. To illustrate this point and as a proof of concept, we have developed two transparent counterfactual explanation techniques that outperform opaque methods. This is largely explained by the fact that opaque approaches often generate non-intuitive counterfactual explanations, \textit{i.e.}, counterexamples that bear no resemblance to the target text. This approach goes against not only the nature of counterfactual explanations but also raises questions about the true level of transparency achieved when explaining one black box with another.

Thus, the key contributions of this paper are as follows:
\begin{enumerate}
    \item The proposal of a spectrum that evaluates the complexity of counterfactual explanations, and provides a nuanced perspective on these methods.
    \item A comparative study of different counterfactual methods, each representing a part of the spectrum.
\end{enumerate}

The document is structured as follows. Section~\ref{sec:trans_opaque} defines transparent and opaque counterfactual explanation methods. Then, Section~\ref{sec:rw} reviews the state of the art. Section~\ref{sec:counterfactual_text} introduces two new transparent methods, which we then analyze in light of the spectrum of existing transparent and opaque techniques (Section~\ref{sec:spectrum}). We then detail the experimental protocol of our comparative study in Section~\ref{subsec:experimentalinfo}. The results of our experiments are presented in Section~\ref{sec:resultscounterfactual}. Section~\ref{sec:conclusion} concludes the paper with a discussion of our findings and the limitations of our study.

\section{Transparent vs. Opaque Methods}
\label{sec:trans_opaque}
In the introduction we categorized counterfactual explanation techniques as either opaque or transparent. We now define these notions in a formal way.

\subsection{Transparent Methods}
Implicitly, transparent counterfactual methods model a text $x \in X$ of length $d$ with words from a vocabulary $\Sigma$, as a binary sparse matrix of dimension $|\Sigma| \times d$. Here $x_{ij}=1$ means that the $i$-th word of the vocabulary $\Sigma$ occurs at position $j$-th position in $x$. A perturbed text $z$ is then obtained as an additive perturbation   
\[ z = X + \epsilon, \;\;\; \text{with} \;\;\;\; z_{ij} = max(0, min(1, x_{ij} + \epsilon_{ij})), \]
where $\epsilon$ is a noise matrix such that $\epsilon_{ij}$ is restricted to three values: $-1$ for removing word $i \in [1,\dots,|\Sigma|]$ at position $j \in [1,\dots,d]$, $0$ for doing nothing, and $1$ for adding word $i$ at position $j$. The clipping operation $\textit{max}(0, \textit{min}(1, \cdot))$ makes sure that $z$ is also a binary matrix.

\subsection{Opaque Methods}
Opaque methods generate counterfactual candidates $z$ by adding noise to the representation of the target text $x \in X$ in a latent space. If we denote such a representation by $g(x)$, this is expressed as \(z = g^{-1}(g(x) + \epsilon)\), where \(g: X \rightarrow \mathbb{R}^{d'} \) is a transformation function into a latent space in $\mathbb{R}^{d'}$ (for some hyperparameter $d'$), and \(\epsilon \in \mathbb{R}^{d'}\) is a noise vector. Opaque methods must also define the inverse function $g^{-1}$ that maps a vector of real numbers into a text.








\section{Related Works}
\label{sec:rw}
Counterfactual explanation methods compute contrastive explanations for ML black-box algorithms by providing examples that resemble a target instance but that lead to a different answer in the black box~\cite{wachter}. These counterfactual explanations convey the minimum changes in the input that would modify a classifier's outcome. Social sciences~\cite{Miller} have shown that human explanations are contrastive and Wachter et al.~\cite{wachter} have illustrated the utility of counterfactual instances in computational law. When it comes to NLP tasks, a good counterfactual explanation should be fluent~\cite{polyjuice}, \textit{i.e.}, read like something someone would say, and be sparse~\cite{counterfactual_review}, \textit{i.e.}, look like the target instance. 

Counterfactual approaches have gained popularity in the last few years. As illustrated by the surveys, first by Bodria et al.~\cite{bodria_benchmark} and later by Guidotti~\cite{survey_cf}, around 50 additional counterfactual methods appeared in a one-year time span. Despite this surge of interest in counterfactual explanations, their study for NLP applications remains underdeveloped~\cite{mice}. 
In the following, we elaborate on the existing counterfactual explanation methods for textual data along a spectrum that spans from transparent to opaque approaches.

\noindent \textbf{Transparent Approaches.} 
One of the first transparent counterfactual explanation approaches was proposed by Martens and Provost~\cite{martens} who introduced Search for Explanations for Document Classification (SEDC), a method that removes the words for which the classifier exhibits the highest \emph{sensitivity}. These are words that impact the classifier's prediction the most. 
More recently, Ross et al.\cite{mice} developed Minimal Contrastive Editing (MICE), a method that employs a Text-To-Text Transfer Transformer to fill masked sentences. In ~\cite{plausible_counterfactual}, the authors presented Plausible Counterfactual Instances Generation (PCIG), which generates grammatically plausible counterfactuals through edits of single words with lexicons manually selected from the economics domain. 

\noindent \textbf{Opaque Methods.} 
Methods such as Decision Boundary~\cite{decision_boundary}, xSPELLS~\cite{xspells} or counterfactualGAN (cfGAN)~\cite{cfgan} operate in three phases. First, they embed the target text onto a latent space. This is accomplished by employing specific techniques such as Variational AutoEncoder (VAE) in the case of xSPELLS, or a pre-trained language model (LM) for cfGAN. Second, while the classifier's decision boundary is not traversed, these methods perturb the latent representation of the target phrase. This is done by adding Gaussian noise in the case of xSPELLS, whereas cfGAN resorts to a Conditional Generative Adversarial Network. Finally, a decoding stage produces sentences from the latent representation of the perturbed documents. 

There also exist methods such as Polyjuice~\cite{polyjuice}, Generate Your Counterfactuals (GYC)~\cite{gyc} and Tailor~\cite{tailor} that perturb text documents in a latent space,  
but can be instructed to change particular linguistic aspects of the target text, such as locality or grammar tense. Such methods are not particularly designed to compute counterfactual explanations but are rather conceived for other applications such as data augmentation.

Unlike pure word-based perturbation methods, latent representations are good at preserving \emph{semantic closeness} for small perturbations. That said, these methods are not free of pitfalls. First, methods such as xSPELLS and cfGAN are deemed opaque since a latent space is not human-understandable~\cite{interpret_gan}. Moreover, existing latent-based approaches do not seem optimized for sparse counterfactual explanations -- one of the defining features of a counterfactual. We show this through our experimental results that suggest that a minor alteration in the latent space can cause a significant alteration in the original space. 

\begin{figure}[t]
    \centering
    \includegraphics[width=\textwidth]{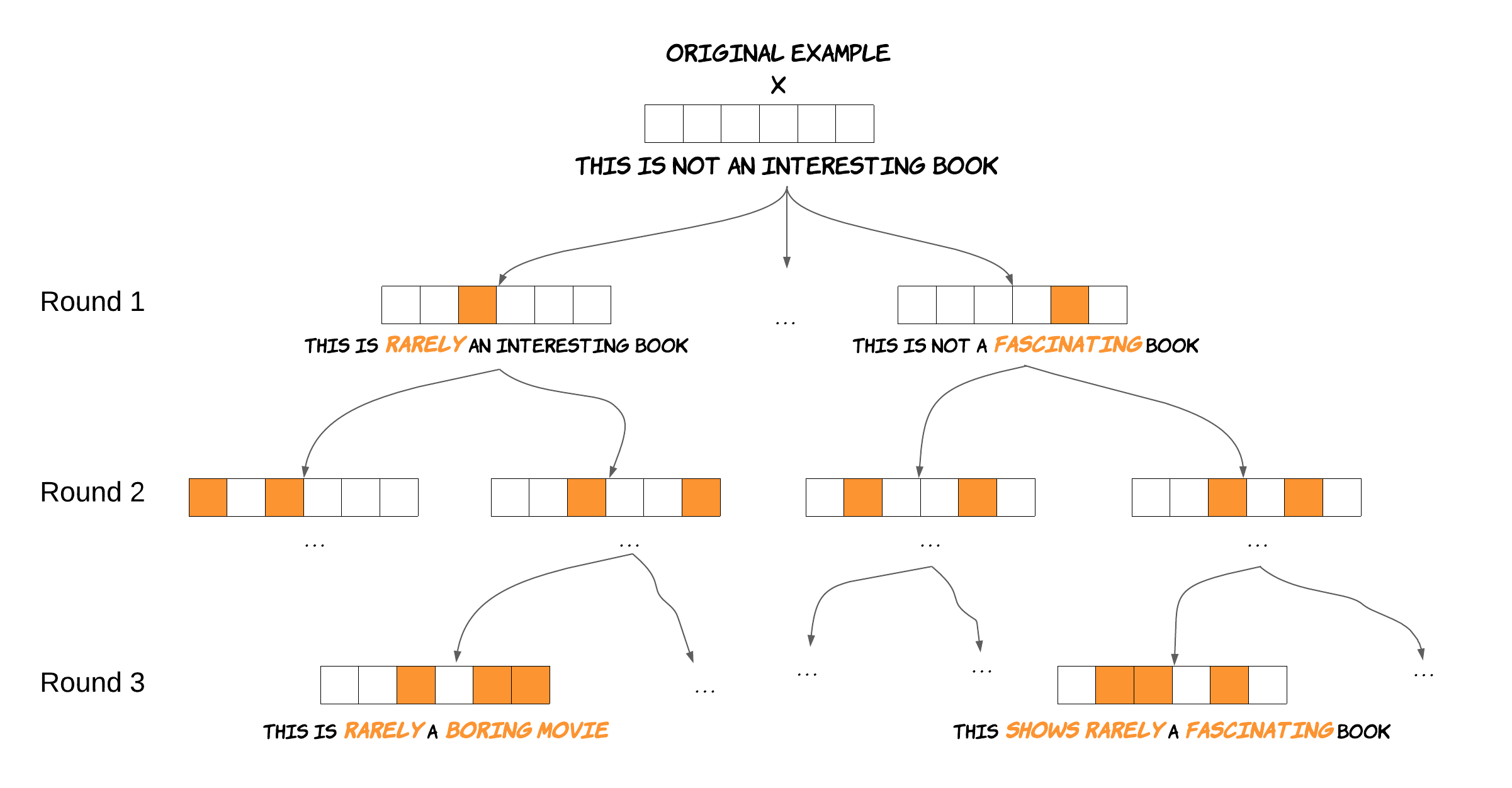}
    \caption{The tree structure of the algorithm used to iteratively perturb the target document. At each round, a word from the target text is iteratively replaced by a word from its corresponding set of potential replacement words. Thus, with each successive round, the number of word replacements for generating artificial documents increases.}
    \label{fig:bfs_growing}
\end{figure}

\section{Counterfactual Explanation Methods for Textual Data}
\label{sec:counterfactual_text}
Before elaborating on our study, we introduce two novel counterfactual explanation techniques, aimed to enrich the middle ground between fully opaque and fully transparent approaches. The methods are called Growing Language and Growing Net, and both depend on an iterative process that replaces words within a target text $x = (x_1, \dots, x_d) \in X$ ($x_i \in \Sigma$ are words from a vocabulary $\Sigma$) until the predicted class of a given classifier $f: X \rightarrow Y$ changes. The goal of such a procedure is to compute sparse counterfactual explanations with the fewest modified words.

\begin{algorithm}
    \caption{Explore}
    \label{alg: growing}
    \begin{algorithmic}[1]
    \Require $\text{target text}\;x = ( x_1, \dots, x_d ) \in X$,
    \Statex $\text{a classifier}\;f: X \rightarrow Y$, 
    \Statex $\textsc{simwords}(\cdot) \rightarrow$ retrieves similar words
    \Statex Hyper-parameters: $n = 2000$
    \Ensure one or multiple counterfactual instances
    \State Initialise $W = (W_1, \dots, W_d)$, sets of candidate words
    \For{$i \gets 1$ \textbf{to} $d$}
        \State $W_i \gets \textsc{simwords}(x_i, \textit{POS}(x_i))$
    \EndFor
    \State Initialize $Z = (z_1, \dots, z_n)$ as $n$ copies of $x$
    \State Initialize $C \leftarrow \emptyset$; $r \gets 0$
    \While{$r < d \land C = \emptyset$}
        \State $r \gets r + 1$
        \For{$j \gets 1$ \textbf{to} $n$} \Comment{For each copy of $x$}
            \For{$l \gets 1$ \textbf{to} $r$}
               \State $k \gets \textit{random}(0, d)$ \Comment{${k : z^k_j = x_k}$} 
                \State $z^k_{j} \gets \text{random word from } W_k$ 
            \EndFor
            \If{$f(x) \neq f(z_j)$}
                \State $C \gets  C \cup \{ z_j \}$
            \EndIf
            
        \EndFor
    \EndWhile
    \State \textbf{return} $C$ 
\end{algorithmic}
\end{algorithm}

Algorithm~\ref{alg: growing} outlines the iterative exploration process employed by Growing Language and Growing Net. In the first step (lines 1 to 4), both approaches generate $d$ sets of potential word replacements  $W_1, \dots, W_d$ for each word $x_i$ in the target document $x$. Those replacements must have the same part-of-speech (POS) tag as $x_i$. The external module to obtain those word replacements depends on the method. These modules are detailed later. Subsequently, our methods create artificial documents iteratively (lines 7 and 18) and exemplified as a tree structure in Figure~\ref{fig:bfs_growing}. These documents are generated while some words in the original document remain non-replaced ($r < d$), or while we have not found any counterfactuals ($C = \emptyset$). At each iteration, the exploration keeps $n$ copies of the original text ($x$) on which we replace $r$ individual words ($x_k$) with randomly selected words from their respective sets of potential replacements ($W_k$). Line 11 ensures that the replaced word comes from the original sentence (${k : z^k_j = x_k}$) and actually replaces $r$ words instead of words already modified. Finally, lines 14-16 check if the resulting phrases are counterfactual instances. 

For example, consider the target review, \textit{``This is not an interesting book''}, classified as negative by a sentiment analysis model (Figure~\ref{fig:bfs_growing}). In the first round, our routine produces artificial reviews with only one modified word. Subsequent rounds will replace two words and so on (lines 10 to 13). In this process, counterfactuals are identified, and the closest one is returned as the explanation. These methods prioritize counterfactuals closely related to the original document to provide concise and meaningful explanations.

\begin{figure}[t]
    \centering
    \includegraphics[width=0.85\textwidth]{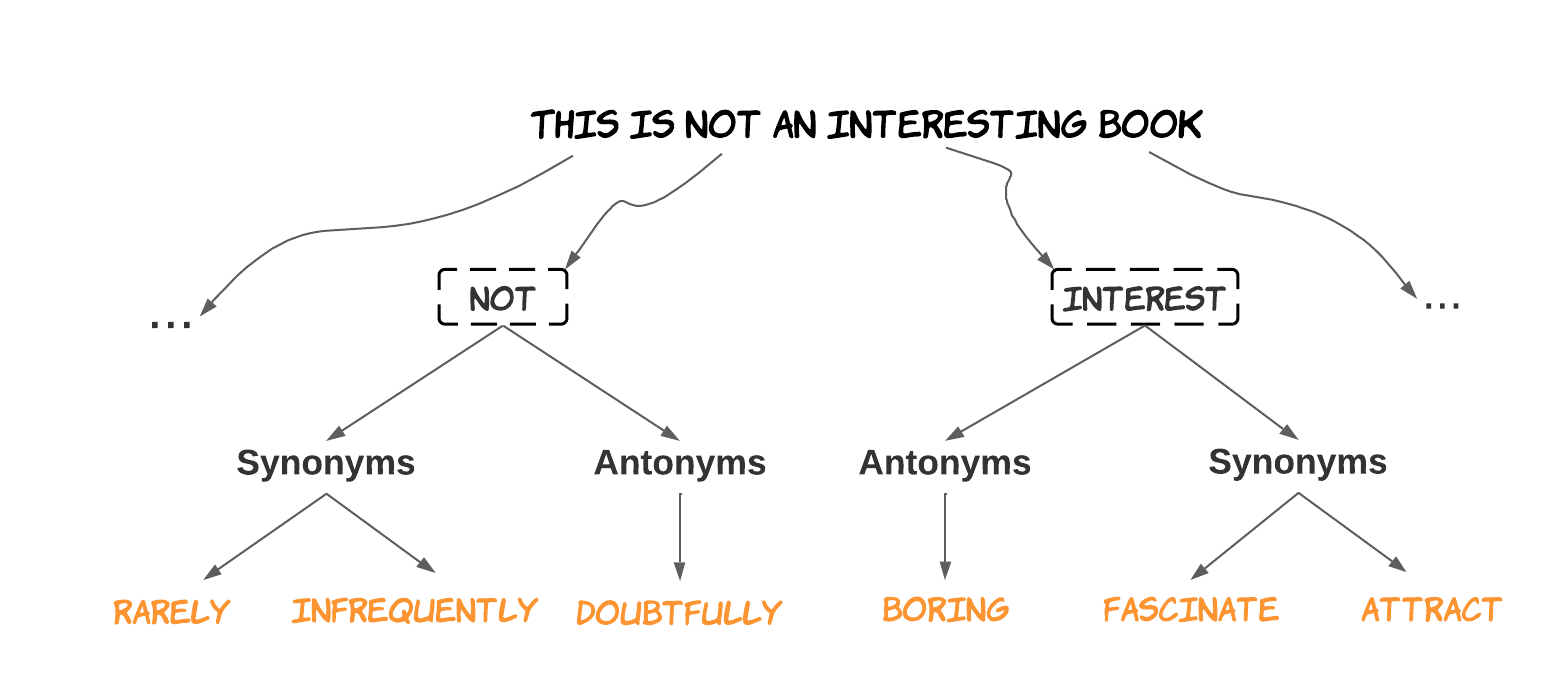}
    \caption{Diagram representing the mechanisms of the Growing Net approach. By leveraging the tree-like structure of WordNet, Growing Net generates sets of words that can replace each term in the target document.}
    \label{fig: growing_net}
\end{figure}
\begin{algorithm} [t]
    \caption{Growing Net}
    \label{alg:growing_net}
    \begin{algorithmic}[1]
    \Require $\text{a target text}\;x = ( x_1, \dots, x_d ) \in X$,  
    \Statex $\text{a classifier}\;f: X \rightarrow Y$;
    \State $C \gets \textit{explore}(x, f, \textsc{wn\_simwords}_{t=1}(\cdot))$
    \State \textbf{return} $\textit{argmax}_{c \in C}{\textit{Wu-P}(c, x)}$
\end{algorithmic}
\end{algorithm}

\subsection{Growing Net}
\label{subsec:growing_net}
This method capitalizes on the rich structure of WordNet~\cite{wordnet} to identify potential word replacements. WordNet is a lexical database and thesaurus that organizes words and their meanings into a semantic tree of interrelated concepts. The method is described in Algorithm~\ref{alg:growing_net}, and uses the module \textsc{wn\_simwords}$_{t}$. In the exploration phase, Growing Net uses \textsc{wn\_simwords}$_{t}$ to find words at a distance of at most $t$ in the WordNet hierarchy among synonyms, antonyms, hyponyms, and hypernyms for a given word $x_i$ to replace. This process is illustrated in Figure~\ref{fig: growing_net}. In our experiments we set $t=1$ as this value already yields good results -- higher values would incur longer runtimes. The exploration returns a set of counterfactuals, from which Growing Net selects the one with the highest Wu-Palmer Similarity (Wu-P)~\cite{wup} as final explanation. This similarity score for text relies on Wordnet, and takes into account the relatedness of the concepts in the phrase, e.g., via the path length to their most common ancestor in the hierarchy. 

\begin{figure}[t]
    \centering
    \includegraphics[width=0.85\textwidth]{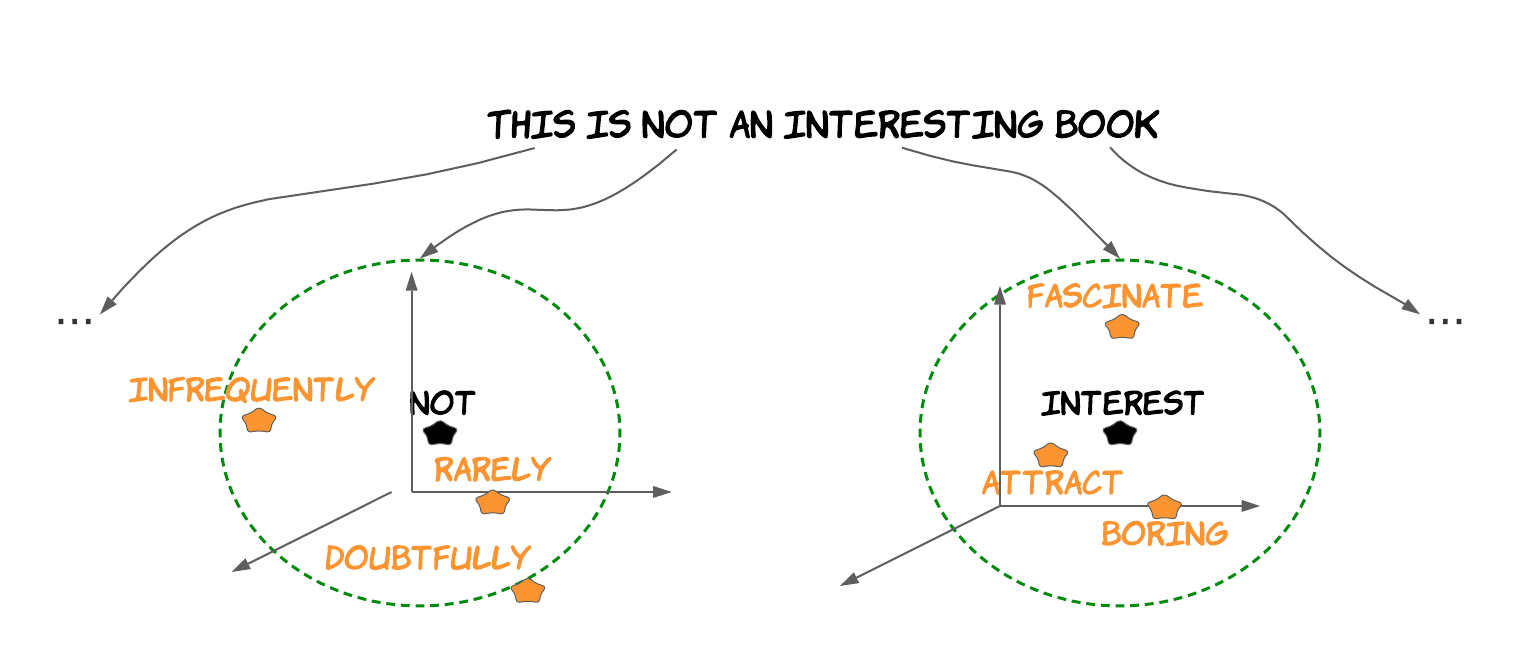}
    \caption{Diagram illustrating the operation of the Growing Language method. The words in the target text are transformed into a latent representation using a large language model. In this latent space, words with similarities become potential replacements for generating artificial documents.}
    \label{fig: growing_language}
\end{figure}
\begin{algorithm}[t]
    \caption{Growing Language}
    \label{alg:growing_language}
    \begin{algorithmic}[1]
    \Require $\text{target text}\;x = ( x_1, \dots, x_d ) \in X$
    \Statex $\text{a classifier}\;f: X \rightarrow Y$;
    \Statex Hyper-parameters: $\tau = 0.02$; $\theta = 0.9$; $\theta_{\textit{min}} = 0.4$;
    \State $C \gets \emptyset$
    \While{$\theta > \theta_{\textit{min}} \land C = \emptyset$ }
    \State $C \gets C \cup \textit{explore}(x, f, \textsc{lm\_simwords}_{\theta}(\cdot))$ 
    \State $\theta \gets \theta - \tau$ 
    \EndWhile
    \State \textbf{return} $\textit{argmin}_{c \in C} ||x - c||_0$
\end{algorithmic}
\end{algorithm}
\subsection{Growing Language}
\label{subsec: growing_language}
This approach leverages the power of language models (LM) to restrict the space of possible word replacements via the module $\textsc{lm\_simwords}_{\theta}$ (see Algorithm~\ref{alg:growing_language}). Large language models are powerful natural language processing AI systems. They are employed in this context to embed words into numerical representations, often referred to as a latent space. In simpler terms, a latent representation consists of high-dimensional vectors that capture the underlying semantic and contextual information of the original text. Given a word $x_i$ to replace, $\textsc{lm\_simwords}_{\theta}$ embeds the word onto the latent space of an LM, as illustrated in Figure~\ref{fig: growing_language}. Then $\textsc{lm\_simwords}_{\theta}$ retrieves words whose latent representation is at a distance of $\theta$ at most. In our experiments, we initially set this threshold to 0.8 on a scale from 0 to 1. If for a given $\theta$, Growing Language cannot find counterfactual instances, the distance threshold is relaxed, i.e., reduced by $\tau$ (set to 0.02 in our experiments), so that the exploration routine considers more words (line 4 Algorithm~\ref{alg:growing_language}). This allows Growing Language to maintain computational efficiency, as starting with a low threshold could result in longer processing times. Should multiple counterfactuals be found, Growing Language selects the one with the fewest modifications compared to the original document (minimal L0 distance). For our experiments, we employed the model en\_core\_web\_md from the library Spacy~\cite{spacy}, but any language model capable of embedding words and offering word distances could be applied in this context.

\section{Interpretability Spectrum}
\label{sec:spectrum}
\begin{figure*}[t]
    \centering
    \includegraphics[width=\textwidth]{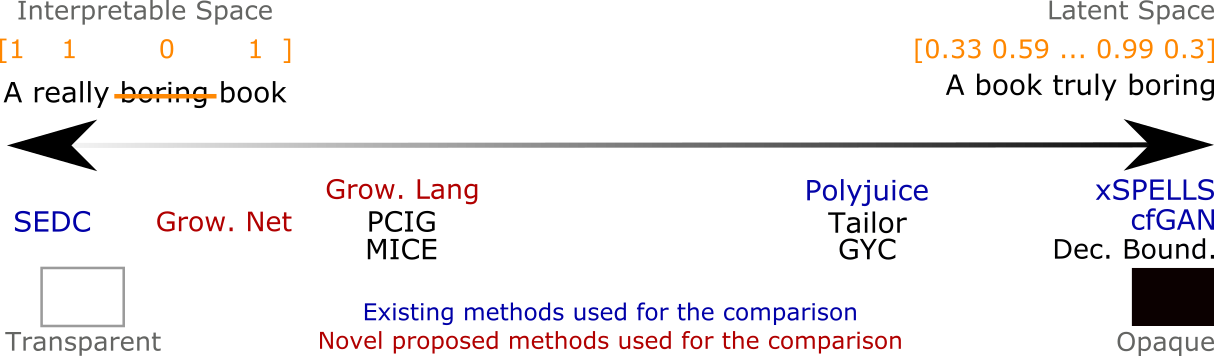}
    \caption{Spectrum for counterfactual explanation techniques that goes from the most transparent methods on the left to the most opaque on the right. 
    Transparent methods perturb documents in a binary space; opaque methods do it in a latent space.}
    \label{fig:complexity_scale}
\end{figure*}

We highlight that the categorisation transparent vs. opaque for counterfactual explanation method defines the two ends of a continuum, which we depict   
in Figure~\ref{fig:complexity_scale}. This spectrum ranges from the most transparent methods on the left to the most opaque methods on the right. We distinguish two sub-categories of transparent methods: \textbf{fully transparent} methods and \textbf{partially transparent} methods. In the first category, individuals can understand why adding noise to a word produces a specific result, such as replacing a word with its antonym, for example. In contrast, for partially transparent methods the choice of the perturbation  \(\epsilon\) may resort to black-box machinery -- even though perturbation itself can be modelled a sum of sparse binary matrices. We also define two sub-categories of opaque methods: \textbf{partially opaque} methods and \textbf{fully opaque} methods. In the first category, the perturbation in the latent space is still controlled by an interpretable symbol, e.g., a control code as in~\cite{polyjuice}. Conversely, it is difficult, if not impossible, to understand the semantics of the perturbation in the latent space of fully opaque methods. We situate the different methods in the state of the art across the spectrum shown in Figure~\ref{fig:complexity_scale}.

\noindent\textbf{Fully Transparent.} At the leftmost end of the spectrum, we find the method SEDC~\cite{martens}, which perturbs text instances by hiding only highly sensitive words within the text. We place Growing Net on the right of SEDC, because it goes beyond simple word masking. Instead, it substitutes words judiciously via an external interpretable asset, namely Wordnet.

\noindent\textbf{Partially Transparent.}  
Methods like PCIG~\cite{plausible_counterfactual}, MICE~\cite{mice}, and Growing Language are considered more opaque than Growing Net, because they employ a latent space to identify semantically close word substitutions. Despite this reliance on black-box techniques, we consider them transparent because the search for counterfactuals is still carried out in the space of words.

\noindent\textbf{Partially Opaque.} Polyjuice, Tailor, and GYC fall in the category of partially opaque methods, as they leverage control codes to perturb the target document. Control codes are specific instructions that adapt the perturbation of the target text so that it complies with a specific task, such as translating, summarizing, or changing the tense of a text. While these modifications occur in a latent space, the inclusion of control codes provides some level of clarity regarding why a modification influences the model's prediction.

\noindent\textbf{Fully Opaque.} On the far right of the interpretability spectrum, we encounter fully opaque approaches such as Decision Boundary, xSPELLS and cfGAN. These methods perturb instances in a latent space, making it challenging for users to discern the underlying process of counterfactual generation. 

This interpretability spectrum provides valuable insights into the transparency and opacity of counterfactual explanation methods, allowing for a more nuanced understanding of their capabilities.

\section{Experimental Protocol}
\label{subsec:experimentalinfo}
Having introduced the spectrum of  counterfactual explanation methods across the interpretability axis, we now describe the experimental setup designed to evaluate those methods. The code of the studied methods, the datasets, and the experimental results are available at \repo{}.

\subsection{Methods}
We picked a set of representative domain-agnostic methods from all regions of the spectrum depicted in Figure~\ref{fig:complexity_scale}.  These include SEDC and Growing Net among the fully transparent methods, Growing Language among the partially transparent methods, Polyjuice among the partially opaque methods, and xSPELLS and cfGAN among the group of fully opaque methods.

It is important to note that we excluded the PCIG~\cite{plausible_counterfactual} and MICE~\cite{mice} methods from our further analysis, each being excluded for specific reasons. In the case of PCIG, this method relies on domain-specific rules in economics, limiting its relevance to our diverse datasets. Regarding MICE, it utilizes transformer models to identify semantically relevant word replacements, which is computationally expensive according to the authors. This complexity runs counter to our goal of favoring simpler transparent methods.

Additionally, we deliberately omitted adversarial methods from our analysis. These methods are designed to induce errors in model predictions rather than for explanatory purposes~\cite{text_attack}. Thus, we excluded them to emphasize the fundamental difference in their objective compared to counterfactual explanation methods. We favored approaches focused on understanding model decisions rather than manipulating its predictions. Similarly, we removed the Linguistically-Informed Transformation (LIT)~\cite{LIT}, a method aimed at automatically generating sets of contrasts. LIT aims to produce documents outside the data distribution, making them unrealistic and diverging from our goal of faithful and relevant explanations.

Detailed information regarding the implementation, versions, and hyperparameters of each counterfactual explanation method used in our experiments is available in Appendix~\ref{appendix:implementation}.

\subsection{Tasks \& Datasets} 
We conduct the evaluation on three popular downstream tasks: (a) spam detection in messages, (b) sentiment analysis, and (c) detection of fake news from newspaper headlines. The datasets associated to these tasks consist of two target classes, and contain between 4000 and 10660 textual documents. The average number of words in each document is between 11.8 and 20.8 as reported in Table~\ref{tab:datasets_counterfactual}. Concerning the polarity~\cite{polarity} and spam detection~\cite{dataset_spam} datasets, we downloaded the data from Kaggle. Regarding the fake news detection dataset, we constructed it using real headlines from a dataset~\footnote{\url{https://www.kaggle.com/datasets/rmisra/news-category-dataset}} combined with headlines fabricated from a fake news dataset~\footnote{\url{https://www.kaggle.com/competitions/fake-news/overview}}. This combined dataset is publicly available in our repository \repo{}. 

\begin{table}[t]
    \centering
    \begin{tabular}{lccccccc}
        \multirow{2}{*}{Dataset} &  \multicolumn{3}{c}{Number of Words} & \multirow{2}{*}{Instances} & \multicolumn{3}{c}{Accuracy (\%)}  \\ 
        \cmidrule(r){2-4} \cmidrule(r){6-8}
        & Total & Average & $\sigma$  & & MLP & RF & BERT \\ \toprule
        Fake  &  19419 & 11.8 & 3.2 & 4025 & 84~\% & 84~\% & 91~\% \\
        Polarity  &  11646 & 20.8 & 9.3 & 10660 &  72~\% & 67~\% & 82~\% \\
        Spam  & 15587 & 18.5 & 10.6 & 8559 & 100~\% & 100~\% & 100~\% \\
        \bottomrule \\
    \end{tabular}
    \caption{Information about the experimental datasets. The ``average'' column denotes the average number of words per instance (document). The column ``$\sigma$ represents the standard deviation while the column ``Instances'' shows the number of text documents per dataset.}
    \label{tab:datasets_counterfactual}
\end{table}

\subsection{Black-box Classifiers}
\label{subsec: black-box_counterfactual}
Our evaluation uses two distinct black-box classifiers implemented using the scikit-learn library and already employed in~\cite{xspells}. These black boxes are (i) a Random Forest (RF) consisting of 500 tree estimators, (ii) a multi-layer perceptron (MLP) with token counts as input, and (iii) a classifier based on DistillBERT with a sequence classification head on top\footnote{\url{https://is.gd/zljjJN}}. For the RF and the MLP, we employed both the \emph{token count} and \emph{tf-idf} vectorizers to convert text into proper inputs for the models. 

We used 70\% of the instances for training, and the remaining for testing. We also selected the target instance to explain within this test set. Across all datasets, the average accuracy of these three classifiers ranges from 67\% to 100\%. Detailed results are presented in Table~\ref{tab:datasets_counterfactual}.

All the experiments were run in a server with an Intel(R) Xeon(R) Gold 5220 CPU (2.20GHz, 18 cores, 24MB L3 cache) and 96GB of RAM (DDR4).

\section{Results}
\label{sec:resultscounterfactual}
We now present the results of our evaluation, organized in four rounds of experiments categorized according to two aspects. First, we assess the quality of the produced counterfactual explanations based on two essential criteria: (i) \textbf{minimality}, 
and (ii) \textbf{plausibility}. Second, we evaluate the methods themselves in terms of (iii) \textbf{flip change}, and (iv) \textbf{runtime}. For each evaluated method and black-box classifier, we computed counterfactual explanations for 100 target texts extracted from the test sets of our datasets.

\subsection{Counterfactual Quality} 
A high-quality textual counterfactual explanation tells us what are the most sensitive parts or aspects of the target phrase, that otherwise changed, would lead to a different classification outcome. As we mentioned in Section~\ref{sec:rw}, it follows then that such an explanation must (i) incur minimal changes w.r.t the target phrase (sparse changes), and (ii) be linguistically plausible, i.e., sound like something a person would naturally write or say~\cite{survey_cf}. 


\begin{figure*}[t]
    \centering
    \includegraphics[width=\textwidth]{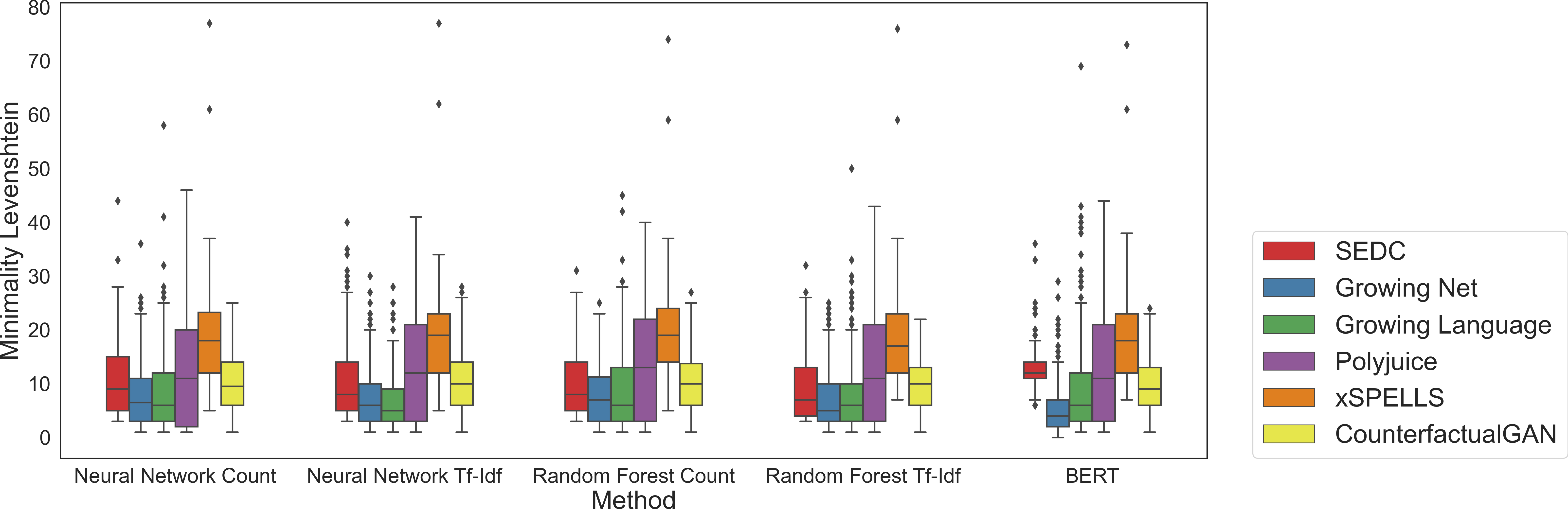}
    \caption{Minimality as the levenshtein edit distance between the closest counterfactual and the target text ($\downarrow$~better).}
    \label{fig:minimality_lev}
\end{figure*}
\begin{figure*}[t]
    \centering
    \includegraphics[width=\textwidth]{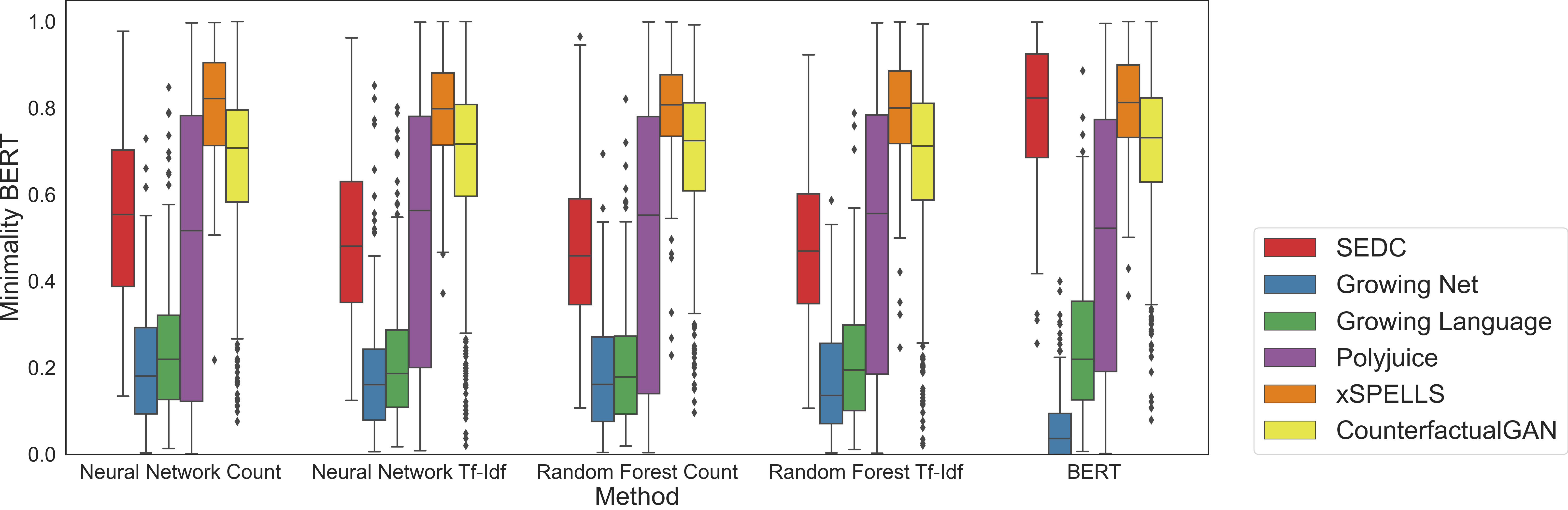}
    \caption{Minimality as the Sentence-BERT embedding distance between the closest counterfactual and the target text ($\downarrow$~better).}
    \label{fig:minimality_bert}
\end{figure*}
\subsubsection{Minimality.} 
We quantify the minimality criterion by measuring the distance between the counterfactual and the target sentence. Figure~\ref{fig:minimality_lev} and \ref{fig:minimality_bert} display the results of our minimality assessments, considering both the Levenshtein distance 
and the cosine similarity within the embedding space of the BERT-Sentence model~\cite{sentence-bert}. This dual approach ensures a comprehensive evaluation, accounting for both lexical similarity and latent features, including aspects of style. 

Notably, our findings reveal that methods positioned in the middle-ground, particularly Growing Net, performed favorably compared to opaque approaches, both in terms of the number of words modified and semantic comparison. It is worth noting that xSPELLS introduced the most significant changes to the original text -- contradicting one of the main functional requirements of a counterfactual explanation~\cite{wachter}. Similarly, we observe a high variance in the minimality of the counterfactuals generated by Polyjuice, indicating that some counterfactuals were notably distant from their corresponding target instances. While these methods introduced minor perturbations to the original text, these modifications occurred within a latent space. Nothing guarantees, however, that these minor adjustments translate into visually subtle modifications of the target phrase when the resulting phrase is brought back to the original space. As an example, consider the target text ``This is one of Polanski's best films.'' from the polarity dataset. For the DistillBERT classifier, cfGAN returns the counterfactual ``this is one of \textbf{shot kingdom intelligence' s all}'', which looks completely unrelated to the target text. Conversely, the transparent method SEDC produces the counterfactual ``This is one of \textbf{MASK MASK MASK}'', whereas Growing Language outputs ``This is one of Polanski's \textbf{worst} films.'' . For more examples of counterfactuals generated by each method, please see appendix~\ref{appendix: illustrative}.

Additionally, we noted first that when the complexity of the classifier increases, the counterfactual explanations generated by SEDC lie farther from the original text. Secondly, we observe minor variations dependent on the vectorizer employed by the classifiers (\textit{count} or \textit{tf-idf}). Hence, for the subsequent phase of the evaluation, we present results exclusively for the tf-idf vectorizer.


\begin{figure*}[t]
    \centering
    \includegraphics[width=\textwidth]{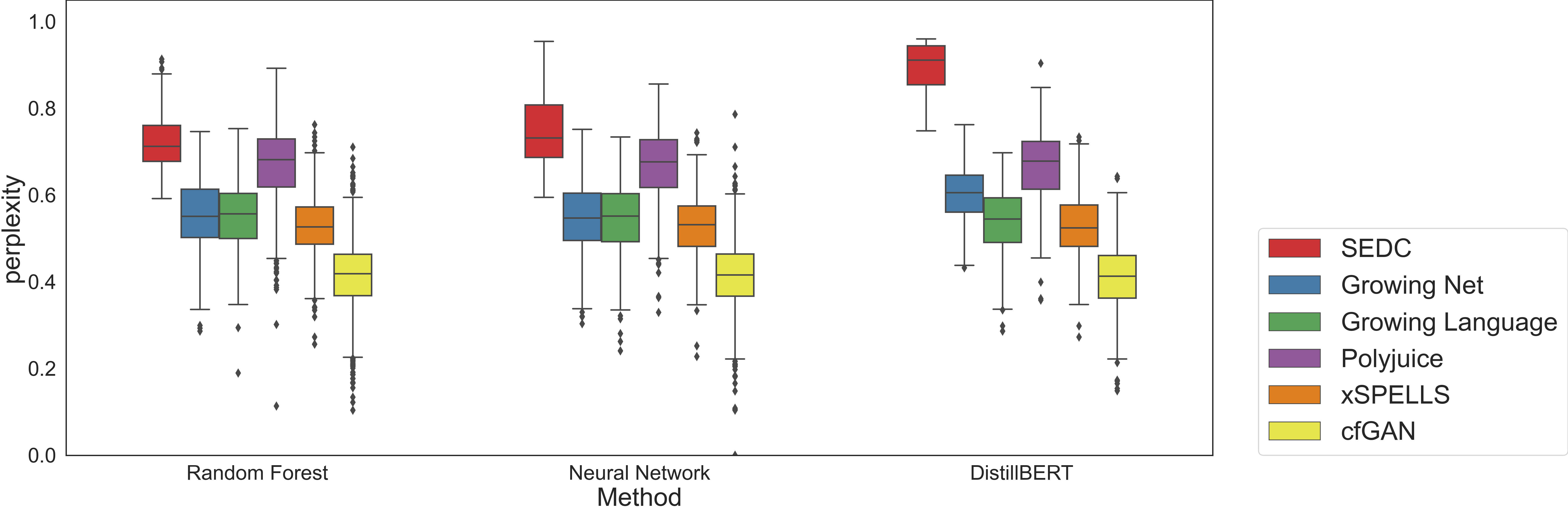}
    \caption{Perplexity as the MSE loss of a GPT model on the generated counterfactuals ($\downarrow$~better).} 
    \label{fig:perplexity}
\end{figure*}

\subsubsection{Plausibility.} 
While linguistic plausibility is typically evaluated through user studies~\cite{gyc, polyjuice}, we approximate it here following the techniques from Ross et al.~\cite{mice,tailor}. Thus, we use perplexity scores based on a GPT language model (GPT-3)~\cite{gpt3}, by calculating the average mean squared error (MSE) loss when predicting the next word in the counterfactual given the preceding words. Figure~\ref{fig:perplexity} presents the plausibility of the counterfactuals. To enhance comparability, we normalized perplexity scores based on the maximum perplexity observed across the entire set of counterfactuals, where lower scores indicate higher plausibility. 
Notably, SEDC and Polyjuice generated texts with the lowest plausibility, which is expected since SEDC masks words, leading sometimes to nonsensical sentences. In contrast, cfGAN demonstrated the highest plausibility, while both Growing Net and Language achieved perplexity scores similar to those of xSPELLS.

\begin{table*}[ht]
    \centering
    \begin{tabular}{lccccccccc}
    \toprule
                     Dataset &    \multicolumn{3}{c}{Fake} &    \multicolumn{3}{c}{Spam}  &    \multicolumn{3}{c}{Polarity} \\ 
                     \cmidrule(r){2-4} \cmidrule(r){5-7} \cmidrule(r){8-10}
                     Black box & MLP & RF & BERT & MLP & RF & BERT & MLP & RF & BERT \\
    \midrule
        \textsc{SEDC}     &  0\textbf{.95} & 0.82 & \textbf{1} & 0.47 & 0.42 & 0.56 & 0.92 & 0.93 & \textbf{0.98} \\
    \midrule
        Grow. Net         & 0.90 & 0.8 & 0.88 & 0.44 & 0.29 & 0.84 & \textbf{0.97} & \textbf{0.98} & 0.90 \\
    \midrule
        Grow. Lang.         & 0.84 & \textbf{0.84} & 0.77 & 0.58 & 0.61 & 0.17 & 0.92 & 0.92 & 0.92 \\
    \midrule
        Polyjuice         & 0.26 & 0.23 & 0.21 & 0.17 & 0.14 & 0.16 & 0.33 & 0.31 & 0.29 \\
    \midrule
        xSPELLS         & 0.68 & 0.78 & 0.77 & \textbf{0.98} & \textbf{0.95} & \textbf{0.91} & 0.91 & 0.76 & 0.91 \\
    \midrule
        cfGAN         & 0.18 & 0.12 & 0.09 & 0.14 & 0.05 & 0.03 & 0.50 & 0.50 & 0.48 \\
    \bottomrule
    \end{tabular}
    \caption{Average label flip per dataset and black box of the six counterfactual methods ($\uparrow$~better).}
    \label{tab:recall}
\end{table*}

\subsection{Method Quality} 
We now compare the quality of the counterfactual explanation methods themselves based on (iii) label flip rate, which measures how frequently a method produces an instance classified differently by the model, and (iv) runtime, the time it takes for each method to generate a counterfactual explanation. 

\subsubsection{Label flip rate.} 
Table~\ref{tab:recall} provides an overview of the label flip results, which indicates the methods' ability to find a counterfactual for a given text document. It is essential to note that, due to the low number of words per dataset (between 11.8 and 20.8), it becomes more challenging for methods to find a counterfactual. Indeed, there are fewer candidate words to modify. We observe that, except for the spam detection dataset, transparent methods achieve the highest label flip rates. This underscores the effectiveness of word replacement with antonyms as a means of discovering counterfactuals. Additionally, xSPELLS demonstrates strong performance for the spam detection dataset and exhibits label flip rates similar to transparent methods for polarity detection.

It is crucial to highlight that the spam detection dataset presents increased difficulty due to the presence of numerous special characters and the informal nature of SMS language. This complexity makes generating counterfactuals more challenging. Furthermore, we emphasize that both Growing Net and Growing Language can be fine-tuned for a more exhaustive search by adjusting their parameters, such as reducing the minimal similarity threshold ($\theta_{\textit{min}}$ in Algorithm~\ref{alg:growing_language}) or exploring more of WordNet's tree structure (increasing $t$ in Algorithm~\ref{alg:growing_net}). While such adjustments may enhance the label flip rate, they may also result in longer runtimes.

\begin{table}[t]
    \centering
    \begin{tabular}{lllll}
    \toprule
    dataset & method & MLP & RF & BERT\\
    \midrule
        \multirow{5}{*}{fake} & SEDC &  31 (14) &  13 (6) & 15 (3)\\
        & Grow. Net &    2 (1) &   1 (1) & 7 (1) \\
        & Grow. Lang. & 55 (28) &     55 (13) & 34 (12)\\
        & Polyjuice & 38 (8) & 70 (185) & 29 (4) \\
        & cfGAN &   1 (0) &    1 (0) & 1 (0) \\
        & xSPELLS &    84 (6) &    86 (7) & 16 (1) \\
        \midrule
        \multirow{5}{*}{spam} & SEDC & 21 (13) &   16 (9) & 16 (6) \\
        & Grow. Net &    1 (1) &    1 (1) & 11 (4) \\
        & Grow. Lang. & 60 (16) &  57 (14) & 88 (43) \\
        & Polyjuice & 32 (7) & 62 (184) & 33 (15) \\ 
        & cfGAN &    1 (0) &   1 (0) & 1 (0) \\
        & xSPELLS &   219 (17) &   198 (16) & 22 (1) \\
        \midrule
        \multirow{5}{*}{polarity} & SEDC &  13 (10) &  12 (9) & 21 (6)\\
        & Grow. Net &   1 (1) &   1 (1) & 9 (2) \\
        & Grow. Lang. &   75 (33) & 74 (32) & 65 (29) \\
        & Polyjuice & 81 (30) & 82 (48) & 29 (4) \\
        & cfGAN &   1 (0) &   1 (0) & 1 (0) \\
        & xSPELLS &   136 (19) &   115 (11) & 24 (2)\\
        \bottomrule
    \end{tabular}
    \caption{Average runtime in seconds of the studied counterfactual methods (and standard deviation).}
    \label{tab:runtime}
\end{table}

\subsubsection{Runtime.} 
Finally, our results regarding execution time are presented in Table~\ref{tab:runtime}. The table details the mean and standard deviation of execution time for each counterfactual explanation method across datasets and classifiers. Notably, cfGAN and Growing Net emerged as the fastest methods for generating counterfactuals. However, it is important to note that cfGAN requires training the Generative Adversarial Network (GAN) on each specific dataset, a process that entails significant training time. The time required for fine-tuning varies, ranging from 4300 seconds for fake news title detection to 6755 seconds for spam detection.

Additionally, we observe that xSPELLS and Growing Language exhibit the slowest performance in terms of execution time. Growing Language, for instance, requires approximately 60 seconds to generate a single counterfactual, while xSPELLS displays execution times ranging from 16 seconds for fake news detection to 219 seconds for spam detection. These findings reveal that, unlike opaque methods like xSPELLS, transparent approaches such as Growing Net are sufficiently fast for real-time explainability.

\section{Discussion \& Conclusion}
\label{sec:conclusion}
Our evaluation provides valuable insights into the landscape of counterfactual explanations for downstream NLP tasks. One of the most striking findings is that complexity, often associated with the use of neural networks and latent spaces, does not necessarily equate to superior performance in this context. Surprisingly, our results demonstrate that simpler approaches, characterized by a systematic and judicious strategy for word replacement, may yield satisfactory outcomes across different quality dimensions. The results of our study prompt 
a deeper reflection on the optimal strategies for generating counterfactual explanations in the field of NLP. 
It invites readers to embrace simplicity and transparency whenever the constraints of the application allow it.

Furthermore, our findings underscore the critical importance of transparency and interpretability in AI and ML,  
especially in high-stakes applications. 
The paradox of explaining a black box with another one  
calls into question the development of opaque approaches when transparent methods suffice, or when transparency is one of the goals in the first place. When focused on NLP applications, our results also call for reflection on the meaning and goal of explanations. If the task is to understand which aspects of a text should change to get a different outcome, a counterfactual explanation that drastically changes every word in the text may not be understandable. On the contrary, a counterfactual based on simple word-masking, albeit simple, may be perceived as implausible. This could hamper the goal of explanations as a means to elicit trust in users.  

That said, the results of our study should be put in context, as the evaluation was conducted on three well-studied downstream applications, namely polarity analysis, fake news detection, and spam detection. Our results might therefore not generalize to other NLP tasks in specialized domains or different languages. While this work puts transparent approaches in the spotlight, our results suggest that plausible counterfactual examples need external domain-adapted knowledge either in the form of language models or knowledge graphs. These may not always be available though. Also, our evaluation was based on popular criteria and  metrics for counterfactual explanations. Specialized applications may still take into account additional criteria such as diversity or actionability. 


Through this work, we expect to encourage the development of more transparent and interpretable AI systems that foster trust and accountability in every step of the AI-driven decision-making processes, either for prediction, recommendation, or explanation.
\bibliographystyle{splncs04}
\bibliography{mybibliography}

\begin{thebibliography}{10}
\providecommand{\url}[1]{\texttt{#1}}
\providecommand{\urlprefix}{URL }
\providecommand{\doi}[1]{https://doi.org/#1}

\bibitem{bodria_benchmark}
Bodria, F., Giannotti, F., Guidotti, R., Naretto, F., Pedreschi, D., Rinzivillo, S.: Benchmarking and survey of explanation methods for black box models. Proc. Data Mining and Knowledge Discovery  (2023). \doi{10.1007/S10618-023-00933-9}, \url{https://doi.org/10.1007/s10618-023-00933-9}

\bibitem{gpt3}
Brown, T.B., Mann, B., Ryder, N., Subbiah, M., Kaplan, J., Dhariwal, P., Neelakantan, A., Shyam, P., Sastry, G., Askell, A., Agarwal, S., Herbert{-}Voss, A., Krueger, G., Henighan, T., Child, R., Ramesh, A., Ziegler, D.M., Wu, J., Winter, C., Hesse, C., Chen, M., Sigler, E., Litwin, M., Gray, S., Chess, B., Clark, J., Berner, C., McCandlish, S., Radford, A., Sutskever, I., Amodei, D.: Language models are few-shot learners. In: Larochelle, H., Ranzato, M., Hadsell, R., Balcan, M., Lin, H. (eds.) Proc. {NeurIPS} (2020), \url{https://proceedings.neurips.cc/paper/2020/hash/1457c0d6bfcb4967418bfb8ac142f64a-Abstract.html}

\bibitem{bert}
Devlin, J., Chang, M., Lee, K., Toutanova, K.: {BERT:} pre-training of deep bidirectional transformers for language understanding. In: Burstein, J., Doran, C., Solorio, T. (eds.) Proc. {NAACL-HLT}. Association for Computational Linguistics (2019). \doi{10.18653/v1/n19-1423}, \url{https://doi.org/10.18653/v1/n19-1423}

\bibitem{wordnet}
Fellbaum, C.: WordNet: An Electronic Lexical Database. Bradford Books (1998)

\bibitem{dataset_spam}
G\'{o}mez~Hidalgo, J.M., Bringas, G.C., S\'{a}nz, E.P., Garc\'{\i}a, F.C.: Content based sms spam filtering. In: Proc. Symposium on Document Engineering. Association for Computing Machinery, New York, NY, USA (2006). \doi{10.1145/1166160.1166191}, \url{https://doi.org/10.1145/1166160.1166191}

\bibitem{survey_cf}
Guidotti, R.: Counterfactual explanations and how to find them: literature review and benchmarking. Data Mining and Knowledge Discovery  (Apr 2022), \url{https://doi.org/10.1007/s10618-022-00831-6}

\bibitem{bias_nlp}
Gururangan, S., Swayamdipta, S., Levy, O., Schwartz, R., Bowman, S.R., Smith, N.A.: Annotation artifacts in natural language inference data. In: Walker, M.A., Ji, H., Stent, A. (eds.) Proc. {NAACL-HLT}. Association for Computational Linguistics (2018). \doi{10.18653/v1/n18-2017}, \url{https://doi.org/10.18653/v1/n18-2017}

\bibitem{decision_boundary}
Hase, P., Bansal, M.: Evaluating explainable {AI:} which algorithmic explanations help users predict model behavior? In: Jurafsky, D., Chai, J., Schluter, N., Tetreault, J.R. (eds.) Proc. {ACL}. Association for Computational Linguistics (2020). \doi{10.18653/v1/2020.acl-main.491}, \url{https://doi.org/10.18653/v1/2020.acl-main.491}

\bibitem{spacy}
Honnibal, M., Montani, I.: {spaCy 2}: Natural lanugage understanding with {B}loom embeddings, convolutional neural networks and incremental parsing (2017), to appear

\bibitem{trends_in_xai}
Jacovi, A.: Trends in explainable {AI} {(XAI)} literature. CoRR  \textbf{abs/2301.05433} (2023). \doi{10.48550/arXiv.2301.05433}, \url{https://doi.org/10.48550/arXiv.2301.05433}

\bibitem{LIT}
Li, C., Shengshuo, L., Liu, Z., Wu, X., Zhou, X., Steinert{-}Threlkeld, S.: Linguistically-informed transformations {(LIT):} {A} method for automatically generating contrast sets. In: Alishahi, A., Belinkov, Y., Chrupala, G., Hupkes, D., Pinter, Y., Sajjad, H. (eds.) Proceedings of the Third BlackboxNLP Workshop on Analyzing and Interpreting Neural Networks for NLP, BlackboxNLP@EMNLP 2020, Online, November 2020. pp. 126--135. Association for Computational Linguistics (2020). \doi{10.18653/v1/2020.blackboxnlp-1.12}, \url{https://doi.org/10.18653/v1/2020.blackboxnlp-1.12}

\bibitem{gan_interpret_latent}
Li, Z., Tao, R., Wang, J., Li, F., Niu, H., Yue, M., Li, B.: Interpreting the latent space of gans via measuring decoupling. {IEEE} Trans. Artif. Intell.  (2021), \url{https://doi.org/10.1109/TAI.2021.3071642}

\bibitem{roberta}
Liu, Y., Ott, M., Goyal, N., Du, J., Joshi, M., Chen, D., Levy, O., Lewis, M., Zettlemoyer, L., Stoyanov, V.: Roberta: {A} robustly optimized {BERT} pretraining approach. CoRR  (2019), \url{http://arxiv.org/abs/1907.11692}

\bibitem{gyc}
Madaan, N., Padhi, I., Panwar, N., Saha, D.: Generate your counterfactuals: Towards controlled counterfactual generation for text. In: Thirty-Fifth Conference on Artificial Intelligence, {AAAI}, Conference on Innovative Applications of Artificial Intelligence, {IAAI}, The Symposium on Educational Advances in Artificial Intelligence, {EAAI}. {AAAI} Press (2021), \url{https://ojs.aaai.org/index.php/AAAI/article/view/17594}

\bibitem{martens}
Martens, D., Provost, F.J.: Explaining data-driven document classifications. {MIS} Q.  (2014), \url{http://misq.org/explaining-data-driven-document-classifications.html}

\bibitem{bias_nlp2}
McCoy, T., Pavlick, E., Linzen, T.: Right for the wrong reasons: Diagnosing syntactic heuristics in natural language inference. In: Korhonen, A., Traum, D.R., Marquez, L. (eds.) Proc. {ACL}. Association for Computational Linguistics (2019). \doi{10.18653/v1/p19-1334}, \url{https://doi.org/10.18653/v1/p19-1334}

\bibitem{Miller}
Miller, T.: Explanation in artificial intelligence: Insights from the social sciences. Artif. Intell.  (2019), \url{https://doi.org/10.1016/j.artint.2018.07.007}

\bibitem{text_attack}
Morris, J.X., Lifland, E., Yoo, J.Y., Grigsby, J., Jin, D., Qi, Y.: Textattack: {A} framework for adversarial attacks, data augmentation, and adversarial training in {NLP}. In: Liu, Q., Schlangen, D. (eds.) Proc. {EMNLP}. Association for Computational Linguistics (2020). \doi{10.18653/v1/2020.emnlp-demos.16}, \url{https://doi.org/10.18653/v1/2020.emnlp-demos.16}

\bibitem{polarity}
Pang, B., Lee, L.: Seeing stars: Exploiting class relationships for sentiment categorization with respect to rating scales. In: Proc. {ACL} (2005)

\bibitem{sentence-bert}
Reimers, N., Gurevych, I.: Sentence-bert: Sentence embeddings using siamese bert-networks. In: Proc. {EMNLP}. Association for Computational Linguistics (2019), \url{http://arxiv.org/abs/1908.10084}

\bibitem{cfgan}
Robeer, M., Bex, F., Feelders, A.: Generating realistic natural language counterfactuals. In: Findings {EMNLP}. Association for Computational Linguistics (2021), \url{https://doi.org/10.18653/v1/2021.findings-emnlp.306}

\bibitem{mice}
Ross, A., Marasovic, A., Peters, M.E.: Explaining {NLP} models via minimal contrastive editing (mice). In: Zong, C., Xia, F., Li, W., Navigli, R. (eds.) Findings {ACL/IJCNLP}. Association for Computational Linguistics (2021). \doi{10.18653/v1/2021.findings-acl.336}, \url{https://doi.org/10.18653/v1/2021.findings-acl.336}

\bibitem{tailor}
Ross, A., Wu, T., Peng, H., Peters, M.E., Gardner, M.: Tailor: Generating and perturbing text with semantic controls. In: Muresan, S., Nakov, P., Villavicencio, A. (eds.) Proc. {ACL}. Association for Computational Linguistics (2022). \doi{10.18653/v1/2022.acl-long.228}, \url{https://doi.org/10.18653/v1/2022.acl-long.228}

\bibitem{xspells}
S.~Punla, C., C.~Farro, R.: Are we there yet?: An analysis of the competencies of {BEED} graduates of {BPSU-DC}. International Multidisciplinary Research Journal  \textbf{4}(3),  50--59 (Sep 2022)

\bibitem{DistilBERT}
Sanh, V., Debut, L., Chaumond, J., Wolf, T.: Distilbert, a distilled version of bert: smaller, faster, cheaper and lighter. ArXiv  \textbf{abs/1910.01108} (2019)

\bibitem{interpret_gan}
Shen, Y., Gu, J., Tang, X., Zhou, B.: Interpreting the latent space of gans for semantic face editing. In: Proc. {CVPR}. Computer Vision Foundation / {IEEE} (2020). \doi{10.1109/CVPR42600.2020.00926}, \url{https://openaccess.thecvf.com/content\_CVPR\_2020/html/Shen\_Interpreting\_the\_Latent\_Space\_of\_GANs\_for\_Semantic\_Face\_Editing\_CVPR\_2020\_paper.html}

\bibitem{counterfactual_review}
Verma, S., Dickerson, J.P., Hines, K.: Counterfactual explanations for machine learning: {A} review. In: NeurIPS 2020 Workshop: ML Retrospectives, Surveys \& Meta-Analyses {ML-RSA}. vol. abs/2010.10596 (2020). \doi{https://arxiv.org/abs/2010.10596}

\bibitem{wachter}
Wachter, S., Mittelstadt, B., Russell, C.: Counterfactual explanations without opening the black box: Automated decisions and the {GDPR}. Harvard Journal of Law and Technology  \textbf{31}(2),  841--87 (2018)

\bibitem{wup}
Wei, X., Ngo, C.: Ontology-enriched semantic space for video search. In: Proc. International Conference on Multimedia. {ACM} (2007), \url{https://doi.org/10.1145/1291233.1291447}

\bibitem{polyjuice}
Wu, T., Ribeiro, M.T., Heer, J., Weld, D.S.: Polyjuice: Generating counterfactuals for explaining, evaluating, and improving models. In: Zong, C., Xia, F., Li, W., Navigli, R. (eds.) Proc. {ACL/IJCNLP}. Association for Computational Linguistics (2021). \doi{10.18653/v1/2021.acl-long.523}, \url{https://doi.org/10.18653/v1/2021.acl-long.523}

\bibitem{plausible_counterfactual}
Yang, L., Kenny, E.M., Ng, T.L.J., Yang, Y., Smyth, B., Dong, R.: Generating plausible counterfactual explanations for deep transformers in financial text classification. In: Proc. {COLING}. International Committee on Computational Linguistics (2020). \doi{https://doi.org/10.18653/v1/2020.coling-main.541}

\end{thebibliography}
\newpage

\appendix
\section{Counterfactual Generation}
\label{appendix:implementation}

We begin by describing the six counterfactual generation methods used to generate counterfactuals. We fill the middle ground with two methods, Growing Net and Growing Language, which implement a strategy similar to existing transparent methods. However, they do so with fewer methodological complexities. We adapted the code used to generate counterfactuals for the three transparent methods (SEDC, Growing Net, and Growing Language) and made it available on GitHub\footnote{\repo{}}. Conversely, we used the original code for the opaque methods, as described below.

\textbf{SEDC}: We modified the code used for word masking to ensure its compatibility with classification models that do not output class probabilities. This modified code version is accessible in Python on our GitHub as a variant of the counterfactual method class. This class proposes to choose from SEDC, Growing Net, or Growing Language, all specialized in generating transparent explanations.

\textbf{Polyjuice:} To generate counterfactuals, we used the code available at the official link \url{https://github.com/tongshuangwu/polyjuice}. We used default hyperparameters with the use of all control codes to perturb texts from each test set until we found 100 instances classified differently by the model.

\textbf{xSPELLS:} We used the V2 version of xSPELLS, available on GitHub \url{https://github.com/lstate/X-SPELLS-V2}, with default hyperparameters.

\textbf{CounterfactualGAN:} We used the code provided in the official release page of the paper, which can be accessed at \url{https://aclanthology.org/2021.findings-emnlp.306/}. We executed CounterfactualGAN (cfGAN) with default hyperparameters.

This comprehensive approach to counterfactual generation ensures a diverse set of methods to be evaluated and compared in our experiments.

\section{Illustrative Example}
\label{appendix: illustrative}
We provide in this section, some examples of counterfactuals generated for each method and each dataset.

\subsection{Fake News Detection}
Original Text: Obama To Apply For Political Asylum In Moneygall

\noindent SEDC: \textbf{MASK MASK MASK} For Political Asylum In Moneygall

\noindent Growing Net : Obama To \textbf{hold} For Political Asylum In Moneygall

\noindent Growing Language : Obama To Apply For \textbf{Hilarious} Asylum In Moneygall

\noindent Polyjuice : Obama is \textbf{expected} To apply for political asylum in \textbf{Guantanamo Bay}

\noindent cfGAN : \textbf{N/A}

\noindent xSPELLS : \textbf{why most states are struggling to}

\subsection{Spam Detection in SMS}
Original Text: Sunshine Quiz Wkly Q! Win a top Sony DVD player if u know which country the Algarve is in? Txt ansr to 82277. aPS1.50 SP:Tyrone

\noindent SEDC: \textbf{MASK MASK} Wkly Q ! Win \textbf{MASK} top \textbf{MASK MASK MASK} if u know \textbf{MASK MASK} the Algarve is in ? \textbf{MASK} ansr \textbf{MASK}. \textbf{MASK} Tyrone

\noindent Growing Net: \textbf{sun test} Wkly Q ! \textbf{bring home} a \textbf{clear} Sony \textbf{videodisc musician} if u know which country the Algarve is in ? Txt ansr to 82277 . aPS1.50 SP : Tyrone

\noindent Growing Language: \textbf{Europe John} Wkly Q ! \textbf{Rest} a \textbf{technical Dr Laptop} player if \textbf{sis} know which country the Algarve \textbf{feels} in ? Txt ansr to 82277 . aPS1.50 \textbf{Gen. – Michigan}

\noindent Polyjuice: \textbf{shine} quiz wkly q! win \textbf{wkly}

\noindent cfGAN: \textbf{\#\#cher week wk mobile two!} win a \textbf{as earthhayaphonic} if u know which country the \textbf{chance week o} is in? \textbf{tt opposed and send fin gives}

\noindent xSPELLS: \textbf{you were your each re not supposed and collect is good way u any please send 50 reply after}

\subsection{Polarity Detection}
Original Text: This is one of Polanski’s best films

\noindent SEDC: This is one of \textbf{MASK MASK MASK}

\noindent Growing Net: This is one of Polanski’s \textbf{ill} films

\noindent Growing Language: This is one of Polanski’s \textbf{worst} films 

\noindent Polyjuice: This is one of Polanski’s \textbf{worst movies}

\noindent cfGAN: This is one of \textbf{shot kingdom intelligence’s all}

\noindent xSPELLS: \textbf{N/A}

\end{document}